\documentclass[smallextended]{svjour3}
\usepackage{amssymb}
\usepackage{amsmath}
\usepackage{graphicx}
\usepackage[numbers]{natbib}
\usepackage{lineno}

\begin{document}
			
\title{Fractal measures of image local features: an application to texture recognition \thanks{J. B. F. gratefully acknowledges the financial support of S\~ao Paulo Research Foundation (FAPESP) (Grant \#2016/16060-0) and from National Council for Scientific and Technological Development, Brazil (CNPq) (Grants \#301480/2016-8 and \#423292/2018-8).}}

\author{Pedro M. Silva \and Joao B. Florindo}

\institute{
	Pedro M. Silva \at
	Federal Institute of Education, Science and Technology of Esp\'{i}rito Santo\\
	Rodovia Governador Jos\'{e} Sete, nº 184 - Itacib\'{a}, CEP 29150-410, Cariacica, ES, Brasil.\\
	\email{icmpedro@gmail.com}
	\and
	Joao B. Florindo \at 	
	Institute of Mathematics, Statistics and Scientific Computing - University of Campinas\\
	Rua S\'{e}rgio Buarque de Holanda, 651, Cidade Universit\'{a}ria "Zeferino Vaz" - Distr. Bar\~{a}o Geraldo, CEP 13083-859, Campinas, SP, Brasil\\
	\email{jbflorindo@ime.unicamp.br}	
}

\date{Received: date / Accepted: date}

\maketitle

\begin{abstract}
Here we propose a new method for the classification of texture images combining fractal measures (fractal dimension, multifractal spectrum and lacunarity) with local binary patterns. More specifically we compute the box counting dimension of the local binary codes thresholded at different levels to compose the feature vector. The proposal is assessed in the classification of three benchmark databases: KTHTIPS-2b, UMD and UIUC as well as in a real-world problem, namely the identification of Brazilian plant species (database 1200Tex) using scanned images of their leaves. The proposed method demonstrated to be competitive with other state-of-the-art solutions reported in the literature. Such results confirmed the potential of combining a powerful local coding description with the multiscale information captured by the fractal dimension for texture classification.
\keywords{Fractal geometry \and Multifractals \and Texture classification \and Box counting \and Local binary patterns.}
\end{abstract}

\section{Introduction}

Texture analysis, and in particular texture recognition, has been one of the most important tasks in computer vision. Despite most of the initial applications of this area being focused on material sciences, during the last decades it has been applied to problems in a wide range of research areas, such as Medicine \cite{KC19}, Biology \cite{VLPMPPGBT18}, Engineering \cite{ZLYF19}, Physics \cite{ZFLW18}, and many others.

Although texture images do not have a consensual formal definition, they are usually associated to structures commonly found in nature and composed by elements whose complexity makes them difficult to be described by the classical Euclidean geometry. This observation naturally opened space for the possibility of analyzing such objects by employing techniques developed in fractal geometry \cite{XJF09,XYLJ10}. Indeed, a fractal is a geometrical set characterized by the self-repetition of basic patterns and by a high degree of complexity. It was in fact described by Mandelbrot in his seminal book \cite{M83} as the ``geometry of nature''. 

Despite the recent popularization of deep learning / neural networks me\-thods in texture analysis, the intrinsic connection between the way that natural structures are composed and the self-similar replication of fractals allows the development of specific purpose models, which present some advantages over deep learning methods, like not requiring so much annotated data for training and the possibility of more straightforward interpretation of the obtained results. The difficulty in obtaining large amounts of annotated data and importance of result interpretation are key aspects, for example, in the analysis of medical images.

However, the theoretical interest in fractals still struggles with some practical limitations. The first and most obvious one is that there is no mathematical fractal in nature. Fractals are defined in infinite scales and this is obviously infeasible in the real world. In this way we need to be aware that the fractal geometry representation is an approximation, a modeling of a real system. A second point is that fractal geometry deals with geometrical sets embedded within a continuous $N$-dimensional space, whereas the most practical way of analyzing a real world object is by inspecting its digital image representation. This issue was addressed for example in \cite{P83}, where a relation between the ``fractality'' of a surface and its respective photography was established. That work also accomplished a psychological experiment that established connections between the fractal dimension and the natural visual perception of attributes such as smoothness and roughness.

Since then, a number of works applying fractal geometry to texture recognition have been presented in the most diverse areas of applications. Nevertheless, a third point still remains without a convincing solution: this is the ideal representation of the image to estimate measures like the fractal dimension. 

Among the attempts we have seen the fractional Brownian motion (fBm) model \cite{ZBM90} (and similar variations like the Fourier-based fractal dimension \cite{R94}) and the geometrical approaches \cite{F04}. fBm has achieved considerable success in texture synthesis, but there is no relevant success reported in recognition. A possible explanation for this is the high degree of specialization of the model, which is the reason for its success in texture synthesis, but at the same time also makes it highly prone to over-fitting in texture recognition. On the other hand, the geometrical approaches rely on interpreting the image as a surface (or cloud of points) embedded in the three-dimensional Euclidean space. This approach has achieved interesting results in texture classification \cite{FCB18}. However, it clearly lacks an effective explanation for using the pixel spatial domain (its localization) and the intensity domain (its gray value) to compose a unique three-dimensional space. Questions like ``how a homogeneous resolution scale can be defined in such conditions?'' are definitely not well answered. 

The multifractal theory \cite{XJF09,XYLJ10,ZHAM19} represented a noticeable attempt to address this issue with an advanced mathematical framework based on Measure Theory. It also achieved promising results on classification problems \cite{XJF09,XYLJ10}, even though not too much of these results have been reported on more modern and challenging texture databases.

Finally, a simple solution would be to take a family of images resulting from thresholding the original texture image at different levels. The direct application of this strategy to the texture image is not feasible as the relation between the gray values of neighbor pixels is lost and this is known to be a fundamental characteristic to distinguish textures of different materials. The authors in \cite{QXSL14} elegantly solved this issue by applying the threshold strategy to the local binary pattern (LBP) \cite{OPM02} map of the texture. This is a representation well known to capture the relations between neighbor pixels \cite{LCS19}. This allows the preservation of such locality information even in the thresholded image.

Based on this context, we propose here the combination of numerical techniques to estimate the fractal dimension of binary images with a well known local encoding of texture images, namely, LBP. We use the fractal dimension of the encoded image thresholded at a range of levels to compose a vector of image descriptors, which are employed for texture classification. To compose the descriptors we verify the use of box-counting and Bouligand-Minkowski fractal dimension as well as the lacunarity measure and the multifractal spectrum. 

The proposed descriptors are compared with other fractal-based texture features, namely, invariant multifractals \cite{XJF09}, wavelet multifractals \cite{XYLJ10}, and pattern lacunarity spectrum \cite{QXSL14}. Other classical and state-of-the-art texture descriptors are also compared, such as VZ-MR8 \cite{VZ05}, local binary patterns \cite{OPM02}, convolutional neural networks \cite{CMKV16}, and others. Comparative tests were carried out over three well known texture benchmark data sets, to know, KTH-TIPS2b, UIUC and UMD. The descriptors were also applied to the identification of species of Brazilian plants (database 1200Tex). The proposed method outperformed the compared approaches in terms of classification accuracy and the results confirmed the potential of such a strategy to provide rich and meaningful texture descriptors. 

\section{Fractal geometry}

Roughly speaking, a fractal is a mathematical object defined at infinite scales and characterized by self-similarity, i.e., repetition of geometrical patterns at different scales and high degree of complexity, which basically means that a hypothetical observer would never see the same object at different scales. It is also commonly distinguished from Euclidean elements for presenting non-integer dimension. Its formal definition is in fact stated in terms of a dimension, technically called Hausdorff dimension.

\subsection{Hausdorff dimension}

Given a set $X$, its Hausdorff measure is defined as: 
\begin{equation}
	\mathcal{H}^s(X) = \lim_{\sigma \rightarrow 0} H^s_{\sigma}(X) \text{ where \ } H^s_{\sigma}(X) = \inf \left\{ \sum_{i=1}^{\infty} |X_i|^s : X_i \mbox{ is a } \sigma\mbox{-cover of } X \right\},
\end{equation}
where $|X_i|$ is the diameter of $X_i$ given by $|X_i| = \sup \{ d(x,y): x,y \in X_i \}$, being $d(x,y)$ a metric. We say that a family of sets $\{ X_i \}$ is a $\sigma$-cover of a set $X$ if:
\begin{equation}
	\left\{
		\begin{array}{l}
			X \subset \bigcup_{i = 1}^{\infty} X_i\\
			0<|X_i|<\sigma.
		\end{array}
	\right.		
\end{equation}

If $X$ is a fractal structure, it can be demonstrated that there exists a real and non-negative value $D$ such that $H^s_{\sigma} = \infty$ for $s<D$ and $H^s_{\sigma} = 0$ for $s>D$. Then $D$ is defined as the \textit{Hausdorff dimension} of $X$. Formally:
\begin{equation}
D = \inf \{ s: \mathcal{H}^s(F)=0 \} = \sup \{ s: \mathcal{H}^s(F)=\infty \}.
\end{equation}

The most widely accepted definition of fractal is that of a geometrical set whose Hausdorff dimension strictly exceeds its Euclidean (topological) dimension.

\subsection{Numerical estimates in fractal geometry}

Although objects with fractal characteristics can be easily found in nature, such real-world structures differ from mathematical fractals in two crucial points: first, they do not have infinite self-similarity; second, the rules dictating the formation of the structure are usually not known. The analytical calculation of the Hausdorff dimension in this scenario is infeasible \cite{M83}. In this context, a large number of numerical values were developed to estimate to which extent these objects can be approximated by a fractal set. Especial interest has been devoted to methods that estimate the fractal dimension of these structures.

Essentially, the computation of the Hausdorff measure involves an infinite covering by units with diameter smaller than $\sigma$. The diameters of these elements are therefore raised to an exponent $s$ and summed up. A discrete approximation of this operation can be accomplished with the aid of an exponential function: 
\begin{equation}
	M_{\sigma} \propto \sigma^s,
\end{equation}
where $M_{\sigma}$  is a measure of the object at the scale $\sigma$, in such a way that any detail larger than $\sigma$ is not counted. By changing the definition of $M$ and $\sigma$, we have definitions of the fractal dimension that are alternative to Hausdorff formulation. These alternative definitions may (and often do) assume values that do not coincide with the Hausdorff dimension, but they preserve the idea of measuring the complexity and spatial occupation of the object: the most complex the structure, the highest the dimension. Many of these alternative definitions are more suitable for numerical computation. Among the most popular definitions possessing this property we have box-counting and Bouligand-Minkowski dimension and lacunarity \cite{F04,YPLL16}.

\subsubsection{Box-counting}

Mathematically, if $X$ is a non-empty bounded subset of $\mathbb{R}^n$ and $N_\delta (X)$ is the smallest number of sets with diameter at most $\delta$ covering $X$, then we define the \textit{lower} and \textit{upper box-counting dimension} of $X$ respectively by
\begin{equation}
	\begin{array}{l}
		\underline{\mbox{dim}}_B(X) = \liminf\limits_{\delta\rightarrow 0}\frac{\log N_\delta (X)}{-\log \delta}\\
		\overline{\mbox{dim}}_B(X) = \limsup\limits_{\delta\rightarrow 0}\frac{\log N_\delta (X)}{-\log \delta}.		
	\end{array}
\end{equation}
If both limits have the same value, we simply call it the \textit{box-counting} dimension:
\begin{equation}\label{eq:box}
	\mbox{dim}_B(X) = \lim\limits_{\delta\rightarrow 0}\frac{\log N_\delta (X)}{-\log \delta}.	
\end{equation}
However, the most interesting equivalent definition of box-counting dimension for practical purposes is that defined under a mesh of hyper-cubes. In $\mathbb{R}^n$ we define a $\delta$-mesh of hypercubes by
\begin{equation}
	[m_1\delta,(m_1+1)\delta] \times \cdots \times [m_n\delta,(m_n+1)\delta],
\end{equation}
where $m_1,\cdots,m_n$ are integer numbers. The measure $N_\delta$ in (\ref{eq:box}) can be replaced by the number of $\delta$-mesh cubes intersecting $X$ as they can be demonstrated to be equivalent measures \cite{F04}.

\subsubsection{Bouligand-Minkowski}

The first step in the calculus of the Bouligand-Minkowski dimension of $X\in\mathbb{R}^n$ is to dilate $X$ by balls in $\mathbb{R}^n$ with radius $\delta$ forming the structure $D(\delta)$ according to:
\begin{equation}
	D(\delta) = \{p:\|p-p'\|_2 \leq \delta, \forall p'\in X\}.
\end{equation}
The hypervolume of $D(\delta)$ is obtained by simply counting the number of points in $D(\delta)$, i.e, 
\begin{equation}
	V(\delta) = \sum_{p\in\mathbb{R}^n} \chi_{D(\delta)}(p),
\end{equation}
where $\chi$ is the characteristic function: $\chi_A(x)=1$ if $x\in A$ and $\chi_A(x)=0$ otherwise.

Finally, the dimension $\mbox{dim}_M$ is provided by
\begin{equation}
	\mbox{dim}_M(X) = n - \lim\limits_{\delta\rightarrow 0}\frac{\log V(\delta)}{\log \delta}.
\end{equation}

\subsubsection{Lacunarity}

For the calculus of the lacunarity of $X$ a hypercube with side-length $\delta$ glides through $X$ and a histogram $H_\delta(k)$ stores the number of hypercubes intersecting exactly $k$ points of $X$. The lacunarity at scale $\delta$ is computed by 
\begin{equation}
		\Lambda_\delta(X) = \frac{E[(H_\delta)^2]}{(E[H_\delta])^2},
\end{equation} 
where $E$ is the expected value for the distribution represented by $H$. Finally, a scale-independent lacunarity is provided by
\begin{equation}
	\Lambda(X) = \lim\limits_{\delta\rightarrow 0}\frac{\log \Lambda_\delta(X)}{\log \delta}.
\end{equation}

\subsubsection{Multifractals}\label{sec:MF}

For the multifractal spectrum we follow the method in \cite{PGBVF01}. First, the image is partitioned into boxes of sizes $L$. Therefore a local measure $\mu_i$ is defined for a range of $q$ values as
\begin{equation}
	\mu_i(q,L) = \frac{\mathfrak{p}_i^q(L)}{\sum_{i=1}^{N(L)}\mathfrak{p}_i^q(L)},
\end{equation}
where $\mathfrak{p}_i(L)$ is the ratio (probability) of white points (points of interest of the analyzed object) falling inside the $i^{th}$ box of size $L$ and $N(L)$ is the number of $L$-sized boxes.

The spectrum $f(q)$ is obtained by
\begin{equation}
	f(q) = \lim\limits_{L\rightarrow 0}\frac{\sum_{i=1}^{N(L)}\mu_i(q,L)\log[\mu_i(q,L)]}{\log L}.
\end{equation}

\section{Proposed method}

We propose here the combined use of box-counting, Bouligand-Minkowski, lacunarity and multifractal spectrum to quantify the fractality of binary images with a well known local encoding of texture images, namely, LBP codes. We use the fractal measure of the encoded image thresholded at a range of levels to compose a vector of image descriptors, which are employed for texture classification.

\begin{figure}
	\includegraphics[width=\textwidth]{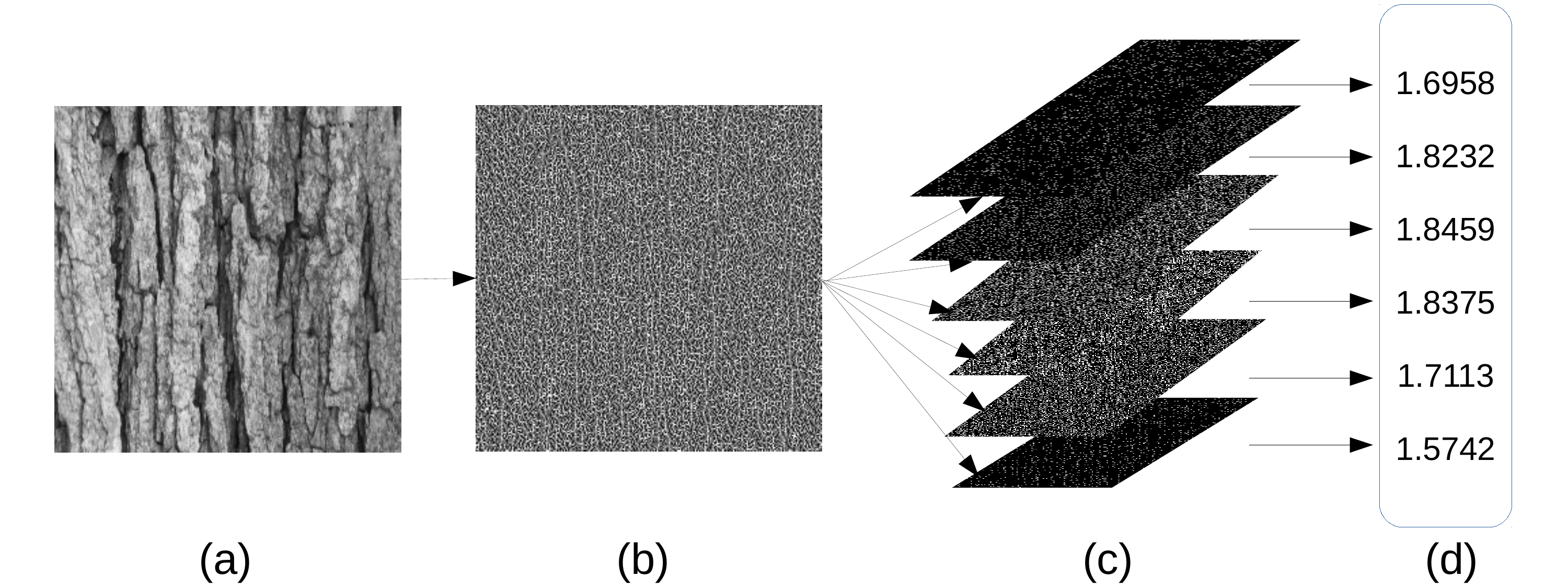}
	\caption{Box counting descriptors for LBP mappings. To simplify the visualization we use $P=4$ and $R=1$ as LBP parameters. (a) Texture. (b) LBP codes. (c) LBP thresholds. (d) Box-counting dimension.}
	\label{fig:methodbc}
\end{figure}

In practice, for a binary image $I$ like that used here, $\delta$ values follow an exponential sequence $\delta=2,4,8,16,\cdots,\log(M)$ where $M$ is the smallest dimension of the image. In the calculus of box-counting dimension, for each $\delta$ the number of squares intersecting the object of interest is accumulated in the variable $N_\delta$ and the numerical dimension is estimated by
\begin{equation}
\mbox{dim}_B(I) = -\alpha,
\end{equation}
where $\alpha$ is the slope of the straight line fit by minimum least squares to the curve $\log \delta \times \log N_\delta$. We also employ the linear coefficient of the straight line to compose the feature vector. Figure \ref{fig:methodbc} illustrates the procedure.

The procedure to compute Bouligand-Minkowski dimension is similar and we use Euclidean distance transform to optimize the calculus of $V(\delta)$ and we employ a maximum radius of $9$, as recommended in \cite{CMB09}. Again, we use the slope and linear coefficient of a straight line fit to $\log \delta \times \log V(\delta)$.

For the lacunarity, the radius $\delta$ ranges between $2$ and $14$, as recommended in \cite{QXSL14} and we also employ both linear and angular coefficients to compose the feature vector.

Finally, the multifractal spectrum is computed according to the description in Section \ref{sec:MF} and using $L = 2,3,5,10,25,50,100,125,250$ as in \cite{PGBVF01}.

\subsection{Motivation}

Fractal dimension and other fractal measures in their strict sense, as in \cite{F04}, are usually not suitable for the analysis of digital images. To start with, digital images are discrete and as a consequence their geometrical representation is countable. Fractal dimension of countable sets is, by definition, zero \cite{F04} and hence such images cannot be identified using fractal dimension. One could construct a continuous model based on the image, but that would introduce artificial data to the process.

Based on this context we develop here a statistical model to investigate how measures like box counting or Bouligand-Minkowski dimension behave in the analysis of digital images. For that purpose, LBP codes can be assumed to follow a uniform distribution, which substantially simplifies the statistical analysis. It is also worth to mention that the other measures explored here, i.e., lacunarity and multifractal spectrum, can be employed in a similar analysis, but the conclusions are expected to be similar, given that both also share the same main objective of measuring the fractality of the image.

\subsubsection{Box counting}

For box counting, we can employ a model similar to that developed in \cite{K13}. Let us suppose that the probability of a white point in the binary image (point of interest) is $p$ and we have a total of $N$ points. Let us also suppose that in $\mathbb{R}^1$ the data is enclosed within the interval $I=[0,1]$. The probability that a box with size $s$ (subinterval of $I$) does not contain any point is
\begin{equation}
	p_0 = (1-s)^{Np}.
\end{equation} 
Correspondingly, the probability of the same box to contain at least one point of interest is the complement of $p_0$:
\begin{equation}
	p_1 = 1 - (1-s)^{Np}.
\end{equation}
The expected number of nonempty boxes used in the calculus of box counting dimension is given by
\begin{equation}
	B(s) = [1 - (1-s)^{Np}]/s.
\end{equation}
Likewise, in $\mathbb{R}^2$ we work on an interval $I^2 = [0,1]\times[0,1]$ and have
\begin{equation}
	B(s) = [1 - (1-s^2)^{Np}]/s^2.
\end{equation}
The trick in \cite{K13} is that for a self-similar set with self-similar dimension $d_S$ we would have
\begin{equation}
	B(s) = [1 - (1-s^{d_S})^{Np}]/s^{d_S}.
\end{equation}
Box counting dimension is obtained by fitting a straight line to the curve $\log s \times \log B(s)$ using least squares. Variable $s$ assumes value within a predefined range (typically linear or exponential) $s_1,s_2,\cdots,s_n$. To simplify the mathematical notation we define $x_i = \log s_i$ and $y_i = \log B(s_i)$. Least squares have a statistical interpretation in which the straight line equation $y = \alpha x + \beta$ is defined in terms of variances and covariances.

\begin{figure}
	\begin{tabular}{ccc}
		\includegraphics[width=.3\textwidth]{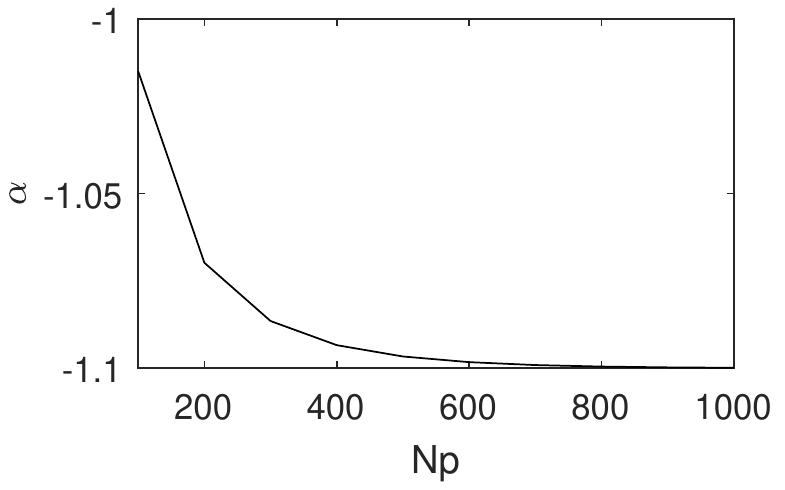}&
		\includegraphics[width=.3\textwidth]{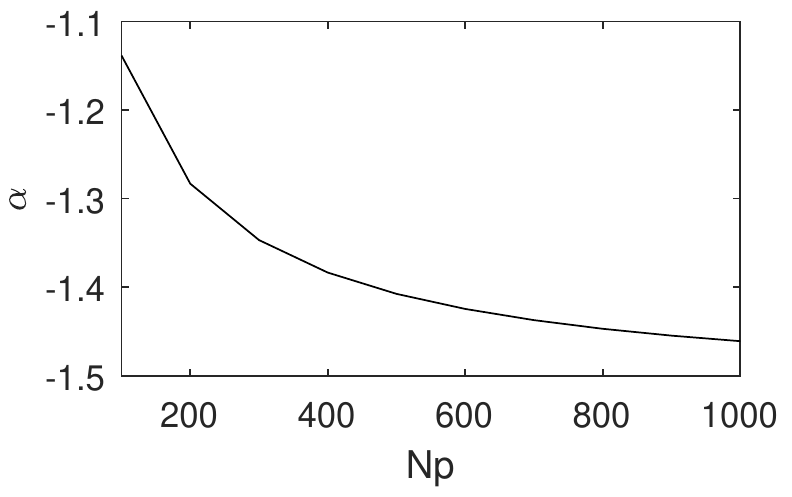}&
		\includegraphics[width=.3\textwidth]{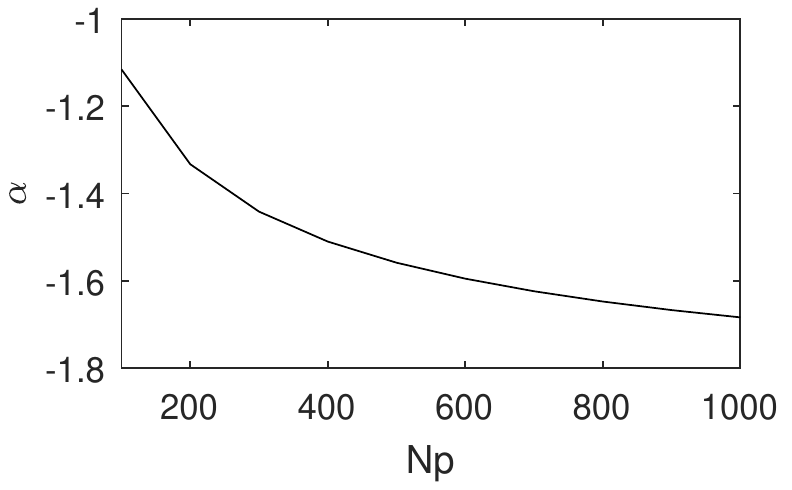}\\
		\includegraphics[width=.3\textwidth]{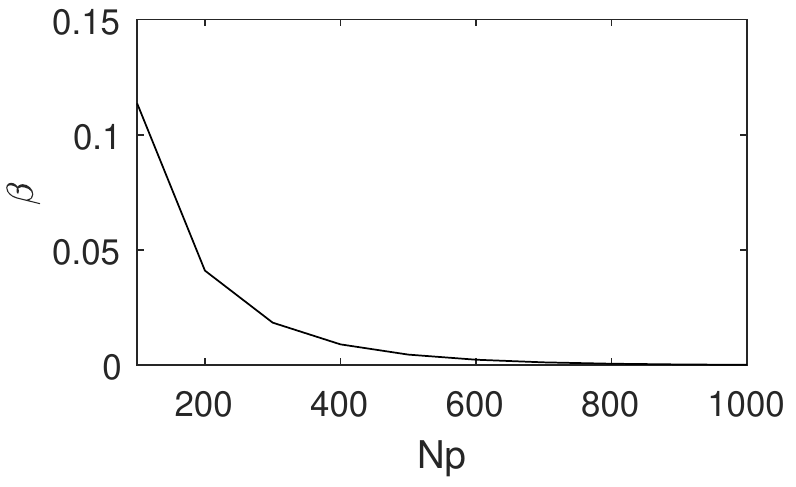}&
		\includegraphics[width=.3\textwidth]{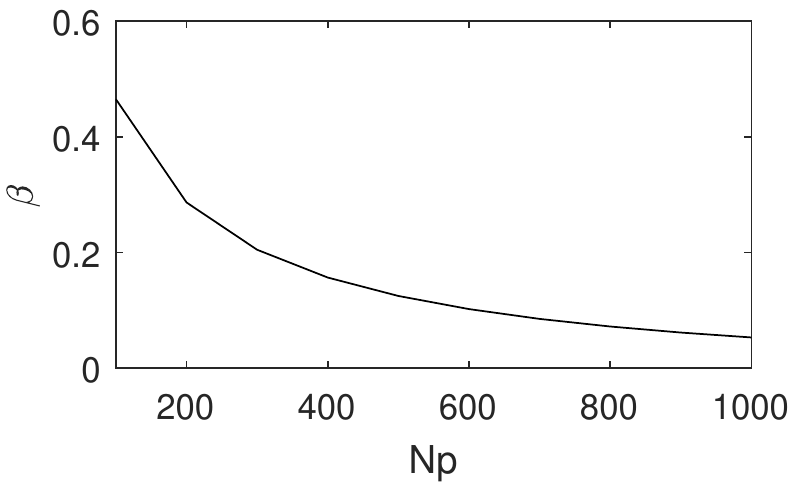}&
		\includegraphics[width=.3\textwidth]{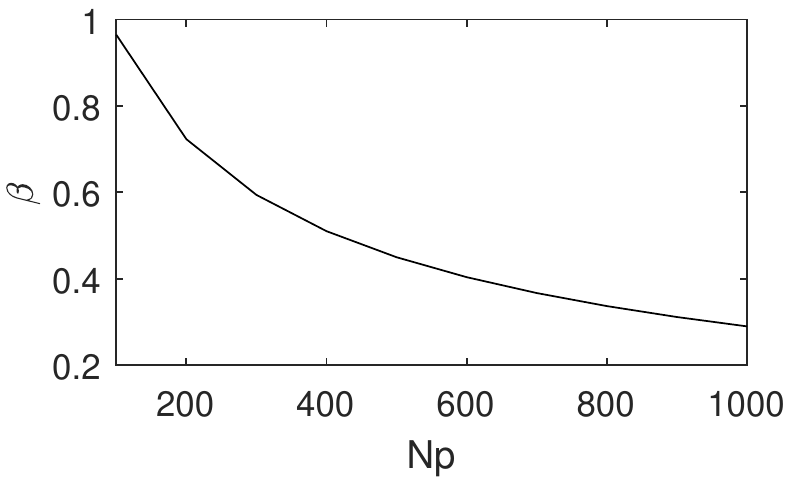}\\	
		$d_S=1.1$ & $d_S=1.5$ & $d_S=1.9$\\	
	\end{tabular}
	\caption{$\alpha$ and $\beta$ in box counting for different numbers of points $Np$ and dimension $d_S$.}
	\label{fig:bc}
\end{figure}

At first, we define the variances of $x$ and $y$ and the covariance between $x$ and $y$:
\begin{equation}
	\begin{array}{ccc}
		\sigma_x^2 = \frac{\sum_{i=1}^{n}(x_i-\overline{x})^2}{n} &
		\sigma_y^2 = \frac{\sum_{i=1}^{n}(y_i-\overline{y})^2}{n} &
		\mbox{cov}(x,y) = \frac{\sum_{i=1}^{n}(x_i-\overline{x})(y_i-\overline{y})}{n},
	\end{array}
\end{equation}
where $\overline{x}$ and $\overline{y}$ denote the mean of $x$ and $y$ respectively. Then it is well known from least squares theory that
\begin{equation}\label{eq:ls}
	\alpha = \frac{\mbox{cov}(x,y)}{\sigma_x^2} \qquad \mbox{and} \qquad \beta = \overline{y} - \alpha\overline{x}.
\end{equation}
Such relations express how the measure (number of boxes) is statistically associated to the scale $s$. Figure \ref{fig:bc} shows $\alpha$ and $\beta$ for different numbers of points $N_p$ and dimension $d_S$. For different similarity dimensions, the box dimension ($\alpha$) and linear coefficient ($\beta$) present different behaviors when plotted against the distribution of points ($Np$). In terms of image descriptors, this corresponds to a more complete representation than the sole dimension or the probability $p$, even though those important features continue to determine the shape of the curve.

\subsection{Bouligand-Minkowski}

A formal analysis of the dilation area of circles is significantly more complicated than for box counting and the interested reader can have some idea of that in \cite{OCFB14}. Nevertheless, for a statistical study, we can reduce the problem to a one-dimensional binary ``image'' in $\mathbb{R}^1$, more specifically over the $[0,1]$ interval. Now the morphological dilation of Bouligand-Minkowski corresponds to placing bars with length $s$ ($s < 1$) centered at points (white points) randomly dropped over the interval. The aim is to compute the expected length covered by at least one bar.

As we need to take care of the boundaries of $[0,1]$, it turns out that for a single bar, the probability $p_1$ of a point at the position $x$ being covered can assume three possible values depending on the value of $x$:
\begin{equation}
	p_1 = 
	\left\{
		\begin{array}{lll}
			x+\frac{s}{2} & \mbox{if} & s \in [0,\frac{r}{2}]\\ 
			s & \mbox{if} & x \in [\frac{s}{2},1-\frac{s}{2}]\\ 
			-x+\frac{s}{2}+1 & \mbox{if} & x \in [1-\frac{s}{2},1].\\ 
		\end{array}
	\right.
\end{equation} 
The probability of $x$ being uncovered is $1-p_1$. In a similar way to what we did in box counting, we suppose that our ``image'' has $N$ points and the probability of occurrence of a white point is $p$. Therefore the expected number of white points is $Np$. The probability of $x$ being uncovered after $n$ bars is randomly placed in $[0,1]$ is $(1-p_1)^{Np}$. Hence the probability of $x$ being covered by at least one bar is
\begin{equation}
	p_n = 1 - (1-p_1)^{Np}.
\end{equation}
Finally the expected length of the covered region is given by
\begin{equation}
	\langle L \rangle = \int_{0}^{1}p_n dx.
\end{equation} 
In practice we work with dilation radii much smaller than the size of the image, which makes possible to disregard the effect of boundaries and focus on the region $x \in [\frac{s}{2},1-\frac{s}{2}]$. This simplifies the integral to
\begin{equation}\label{eq:length_n}
	\langle L \rangle = 1 - (1-s)^{Np}.
\end{equation}
This expression can be rewritten as
\begin{equation}
	\langle L \rangle = 1 - \left( 1-\frac{Nps}{Np} \right)^{Np},
\end{equation}
which for large $N$ allows for the use of the exponential limit:
\begin{equation}
	\lim\limits_{n\rightarrow\infty} \langle L \rangle = 1 - \mathrm{e}^{-Nps}.
\end{equation}
This limit confirms the power law relation naturally appearing in fractal-like phenomenons. However, for our purposes, it is also worthwhile to investigate (\ref{eq:length_n}), which determines the behavior of $\langle L \rangle$ for not so large values of $N$.

\begin{figure}
	\begin{tabular}{ccc}
		\includegraphics[width=.3\textwidth]{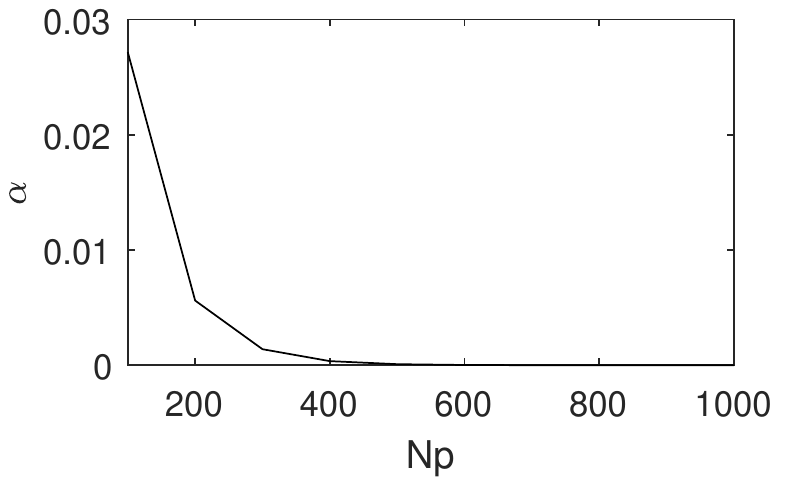}&
		\includegraphics[width=.3\textwidth]{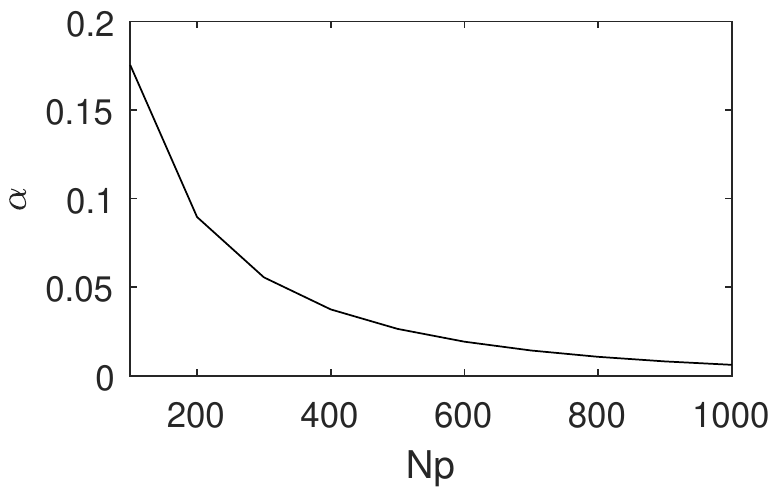}&
		\includegraphics[width=.3\textwidth]{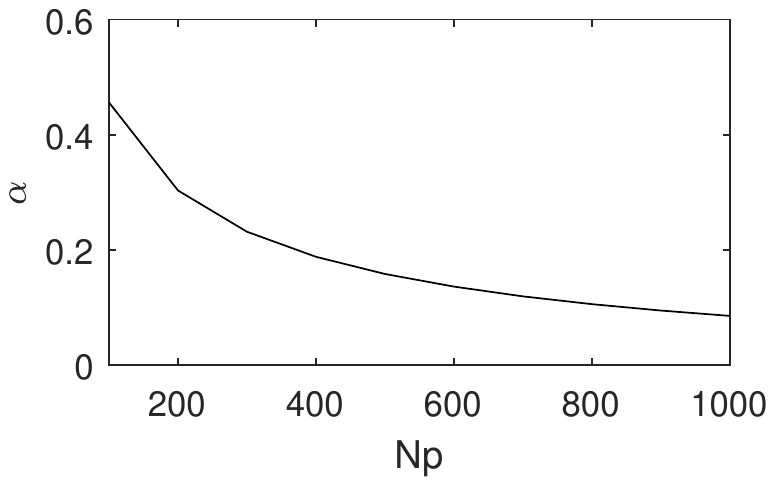}\\
		\includegraphics[width=.3\textwidth]{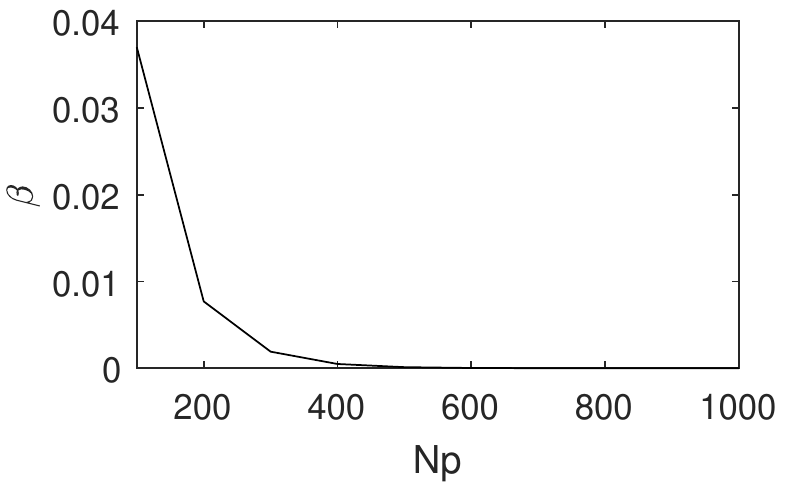}&
		\includegraphics[width=.3\textwidth]{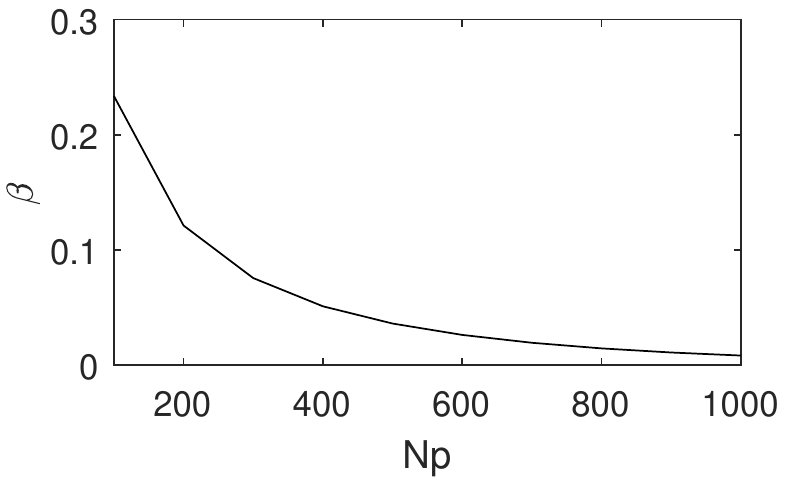}&
		\includegraphics[width=.3\textwidth]{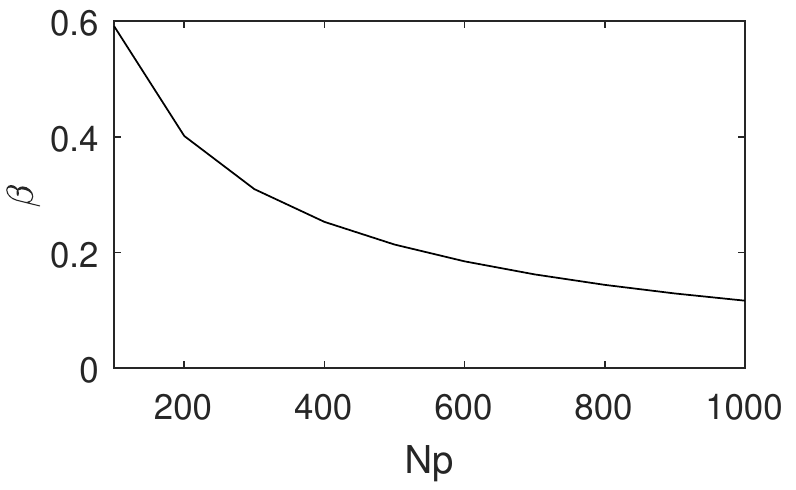}\\	
		$d_S=1.1$ & $d_S=1.5$ & $d_S=1.9$\\	
	\end{tabular}
	\caption{$\alpha$ and $\beta$ in Bouligand-Minkowski for different numbers of points $Np$ and self-similar dimension $d_S$.}
	\label{fig:bm}
\end{figure}

Using a strategy similar to that in \cite{K13}, in two dimensions we have circles with radius $s$ and ($\ref{eq:length_n}$) is rewritten as
\begin{equation}
	\langle L \rangle = 1 - (1-\pi s^2)^{Np}.
\end{equation}
For a self-similar structure with self-similar dimension $d_S$ we can resort to the formula for the volume of the $n$-dimensional ball, yielding
\begin{equation}
	\langle L \rangle = 1 - \left( 1-\frac{\pi^{d_S/2}}{\Gamma\left( \frac{d_S}{2}+1 \right)} s^{d_S} \right)^{Np},
\end{equation}
where $\Gamma$ is the Euler's gamma function. $\alpha$ and $\beta$ are provided by (\ref{eq:ls}) as usual. Figure \ref{fig:bm} shows $\alpha$ and $\beta$ for different numbers of points $N_p$ and self-similar dimension $d_S$. As in Figure \ref{fig:bc} we have different curves for each similarity dimension.

In general, all this analysis corroborates the interest in analyzing local characteristics of the image like LBP codes by inspecting properties of the $\log-\log$ dimension curve. Here we use the slope and linear coefficient of such curve and verify that such potential theoretically predicted is also confirmed in practical situations.

\section{Experiments}

The performance of the proposed descriptors was evaluated on three benchmark databases widely used in the recent literature of texture classification methods. We also applied the same method to a practical problem, namely the identification of species of Brazilian plants using scanned images of leaves.  

The first benchmark data set is KTHTIPS-2b \cite{HCFE04}, a database comprising 4752 images uniformly divided into 11 material classes. An important characteristic of this data is its focus on the material represented in the image rather than on the instance of the photographed object. In each material class the images can still be divided into 4 samples. Each sample follows a particular scheme of scale, pose and illumination. The validation protocol is the most typically employed in the literature, where 1 sample is used for training and the remaining 3 samples are used for testing. The accuracy (percentage of images correctly classified) and respective standard deviation are obtained by averaging out the results for the 4 possible combinations of training/testing.

The second database is UIUC \cite{LSP05}, containing 1000 images evenly divided into 25 texture categories. The images were collected under non-controlled conditions and contain variation in albedo, perspective, illumination and scale. For the validation split we randomly select 20 images of each texture for training and the remaining 20 images for testing. This is repeated 10 times to provide the average accuracy and deviation.

The third data set is UMD \cite{XJF09}. This is composed by a collection of 1000 high-resolution images collected by a family camera without any illumination control. The images are uncalibrated and unregistered and are categorized into 25 classes, each one with 40 images. Each image has a resolution of $1280\times 960$. The database is characterized by high variance in viewpoint and scale, what makes its classification a more challenging task.

To reduce dimensionality and attenuate the effect of redundant information, the proposed descriptors are subject to principal component analysis \cite{J02} before the classification. For the classification, we tested two possibilities: support vector machines (SVM) \cite{CV95} with a configuration similar to what is employed in \cite{CMKV16}, i.e., linear kernel, $C=1$ and $L^2$ normalization, and linear discriminant analysis (LDA) \cite{K88}.

\section{Results and Discussion}

An initial test was accomplished to verify the performance of the proposed method for the two compared classifiers, i.e., LDA and SVM. For this test we employed box counting descriptors as those have more straightforward interpretation and allow more unbiased evaluation of the classifier. Table \ref{tab:classif} lists the percentage of images correctly classified in each compared database as well as the respective standard deviation. LDA provided higher accuracies in all tested data and, based on this, the remaining experiments are carried out using this classification scheme.
\begin{table}
	\centering
	\caption{Classification accuracy using SVM and LDA classifier.}
	\label{tab:classif}
	\begin{tabular}{ccc}
		\hline
			Database & SVM & LDA\\	
		\hline
			KTHTIPS-2b & 63.4$\pm$3.4 & \textbf{65.5$\pm$3.5}\\
			UIUC & 96.3$\pm$0.5 & \textbf{97.3$\pm$0.6}\\
			UMD & 98.0$\pm$0.7 & \textbf{98.6$\pm$0.6}\\
			1200Tex & 81.9$\pm$1.7 & \textbf{86.6$\pm$1.1}\\
		\hline
	\end{tabular}
\end{table}

In Table \ref{tab:indiv} we analyze the performance of the individual fractal descriptors: box counting (BC), Bouligand-Minkowski (BM), lacunarity (L), and multifractal (MF). From that table it is possible to state that each fractal metric can be more or less recommended depending on particularities of the data set. This also suggests that combining different fractal features could provide even better classification results.
\begin{table}
	\centering
	\caption{Classification accuracy of the individual fractal descriptors: BC (box counting), BM (Bouligand-Minkowski), L (lacunarity), and MF (multifractal).}
	\label{tab:indiv}
	\begin{tabular}{ccccc}
		\hline
			Database & BC & BM & L & MF\\	
		\hline
			KTHTIPS-2b & 65.5$\pm$3.5 & \textbf{66.3$\pm$3.1} & 65.9$\pm$3.6 & 58.9$\pm$3.3\\
			UIUC & 97.4$\pm$0.5 & 97.6$\pm$0.4 & \textbf{97.8$\pm$0.5} & 86.7$\pm$1.4\\
			UMD & 98.6$\pm$0.6 & 99.0$\pm$0.6 & \textbf{99.1$\pm$0.4} & 97.2$\pm$0.6\\
			1200Tex & \textbf{86.4$\pm$1.1} & 85.9$\pm$1.3 & 84.9$\pm$1.1 & 67.2$\pm$1.5\\
		\hline
	\end{tabular}
\end{table}

Table \ref{tab:comb} shows the accuracy when some of the investigated fractal features are combined into a single vector of descriptors. Other combinations were veri\-fied but at the end those ones presented in Table \ref{tab:comb} had the most competitive performances. In general, some subtle improvement over the individual features was obtained in most data sets when combined descriptors were used.
\begin{table}
	\centering
	\caption{Classification accuracy for some combinations of fractal descriptors. Best results in each data set are in bold.}
	\label{tab:comb}
	\begin{tabular}{ccccccc}
		\hline
			Database & BC+BM & BC+L & BC+MF & BM+L & BM+MF & L+MF\\	
		\hline
			KTHTIPS-2b & 67.5$\pm$2.1 & \textbf{67.6$\pm$2.1} & 65.0$\pm$2.1 & 67.4$\pm$2.6 & 65.4$\pm$2.2 & 66.7$\pm$3.1\\
			UIUC & 97.9$\pm$0.5 & \textbf{98.1$\pm$0.5} & 96.4$\pm$0.7 & 98.0$\pm$0.4 & 96.3$\pm$0.8 & 97.0$\pm$0.6\\
			UMD & 98.9$\pm$0.6 & 99.1$\pm$0.5 & 99.1$\pm$0.5 & 99.1$\pm$0.4 & \textbf{99.3$\pm$0.2} & 99.1$\pm$0.3\\
			1200Tex & 86.2$\pm$1.0 & 86.3$\pm$0.8 & 82.2$\pm$1.6 & 85.9$\pm$1.1 & 81.6$\pm$1.8 & 80.5$\pm$1.6\\
		\hline
	\end{tabular}\\
	\begin{tabular}{ccc}
		\hline
		Database & BC+BM+L & BC+BM+L+MF\\	
		\hline
		KTHTIPS-2b & 66.4$\pm$2.0 & 66.4$\pm$2.0\\
		UIUC & \textbf{98.1$\pm$0.6} & 97.6$\pm$0.5\\
		UMD  & 99.0$\pm$0.5 & 99.2$\pm$0.3\\
		1200Tex & \textbf{86.3$\pm$0.6} & 82.5$\pm$1.4\\
		\hline
	\end{tabular}
\end{table}

Table \ref{tab:SRdatabase} lists the accuracy performance of the proposed descriptors in KTH\-TIPS-2b, UMD and UIUC, compared with other results published in the literature. Details concerning parameters and other implementation details for each result can be obtained in the respective references. Here the fractal features outperformed advanced approaches like SIFT $+$ VLAD or SIFT $+$ BoVW in KTHTIPS-2b. Methods based on automatic (deep) learning like FC-CNN were also outperformed in UIUC and UMD. These are canonical examples of what could be defined as ``textures in their strict sense'' and the results confirmed the potential of fractal-based methods to analyze such types of images. Finally, MFS and PLS are examples of fractal-based approaches for texture recognition. Both were also outperformed by the proposed descriptors in UIUC and UMD.
\begin{table}[!htpb]
	\centering
	\caption{Accuracy of the proposed descriptors compared with other texture descriptors in the literature. All the results except for the proposed method were obtained from the literature. A `-' indicates that no result was published for that method on that database. A superscript $^1$ denotes a slightly different protocol in KTH-TIPS2b where 3 samples are used for training and 1 sample for testing.}
	\label{tab:SRdatabase}
		\begin{tabular}{lccc}
			Method & KTH-TIPS2b & UIUC & UMD\\
			\hline
			VZ-MR8 \cite{VZ05} & 46.3 & 92.9 & - \\
			LBP \cite{OPM02} & 50.5 & 88.4 & 96.1\\
			VZ-Joint \cite{VZ09} & 53.3 & 78.4 & - \\
			BSIF \cite{KR12} & 54.3 & 73.4 & 96.1\\					
			LBP-FH \cite{AMHP09} & 54.6 & - & - \\			
			CLBP \cite{GZZ10} & 57.3 & 95.7 & 98.6 \\
			SIFT+LLC \cite{CMKV16} & 57.6 & 96.3 & 98.4\\					
			ELBP \cite{LZLKF12} & 58.1 & - & - \\
			SIFT + KCB \cite{CMKMV14} & 58.3 & 91.4 & 98.0\\
			LBP$_{riu2}$/VAR \cite{OPM02} & 58.5$^1$ & 84.4 & 95.9\\			
			SIFT + BoVW \cite{CMKMV14} & 58.4 & 96.1 & 98.1\\
			PCANet (NNC) \cite{CJGLZM15} & 59.4$^1$ & 57.7 & 90.5\\				
			SIFT + VLAD \cite{CMKMV14} & 63.1 & 96.5 & 99.3\\
			RandNet (NNC) \cite{CJGLZM15} & 60.7$^1$ & 56.6 & 90.9\\
			SIFT+IFV \cite{CMKMV14} & 58.2 & 97.0 & 99.2 \\
			ScatNet (NNC) \cite{BM13} & 63.7$^1$ & 88.6 & 93.4\\
			DeCAF \cite{CMKMV14} & 70.7 & 94.2 & 96.4\\
			FC-CNN VGGM \cite{CMKV16} & 71.0 & 94.5 & 97.2\\	
			MFS \cite{XJF09} & - & 92.7 & 93.9 \\							
			PLS \cite{QXSL14} & - & 96.6 & 99.0\\
			\hline
			Proposed & 67.6 & 98.1 & 99.3\\
			\hline		
		\end{tabular}
\end{table}

\begin{figure}
	\begin{tabular}{c}
		\begin{tabular}{cc}
			\includegraphics[width=.45\textwidth]{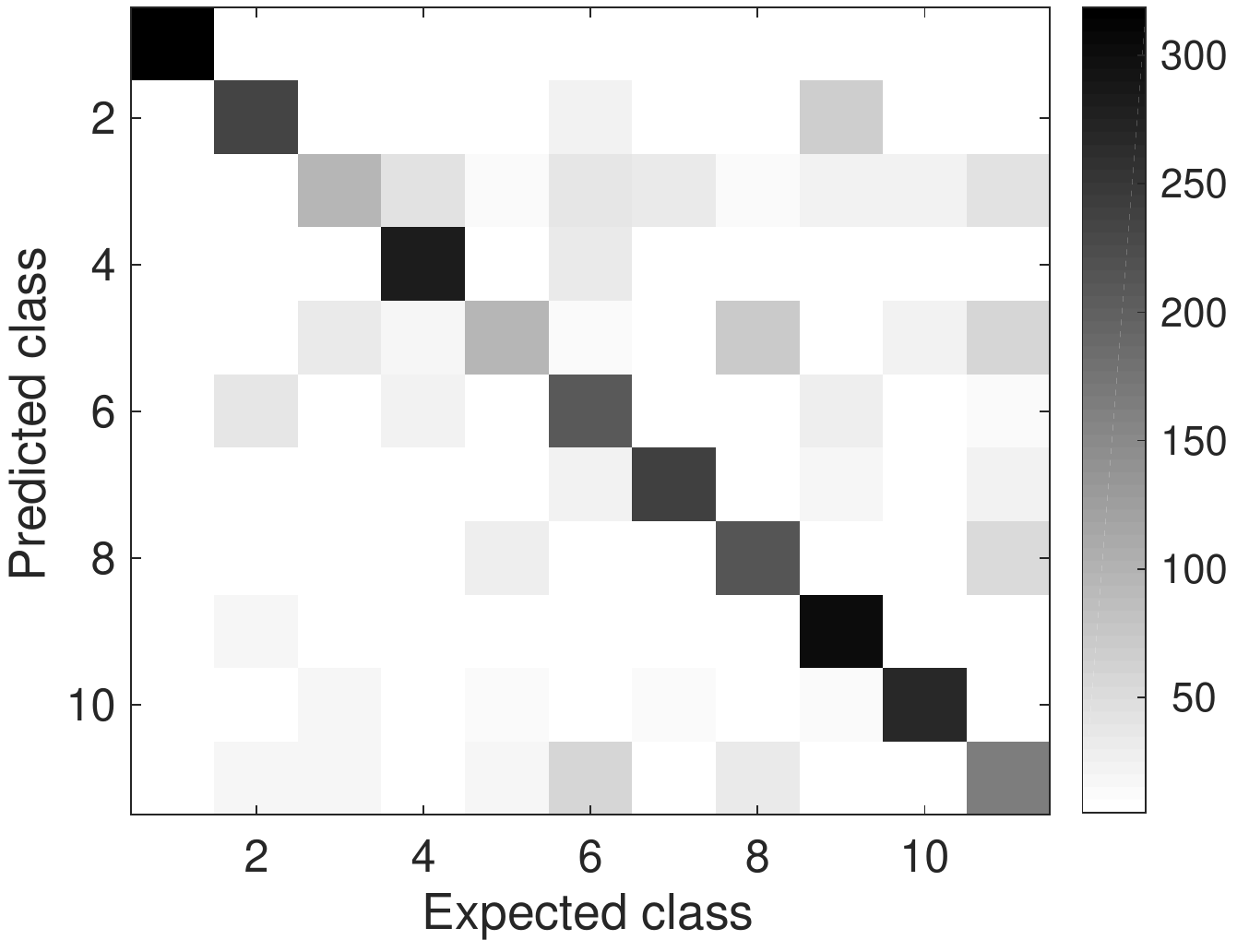} &
			\includegraphics[width=.45\textwidth]{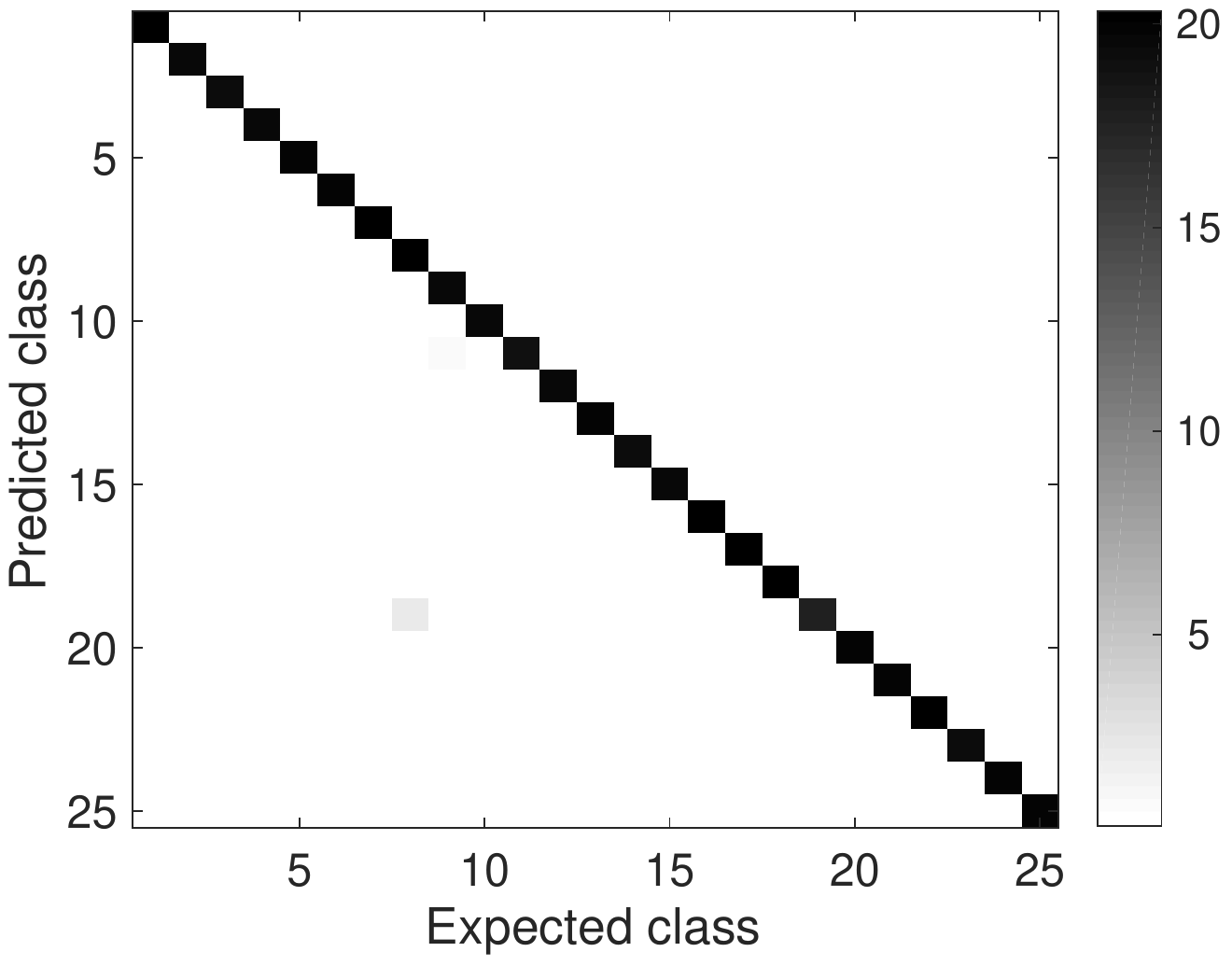}\\
			(a) & (b)\\
		\end{tabular}\\
		\begin{tabular}{c}
			\includegraphics[width=.45\textwidth]{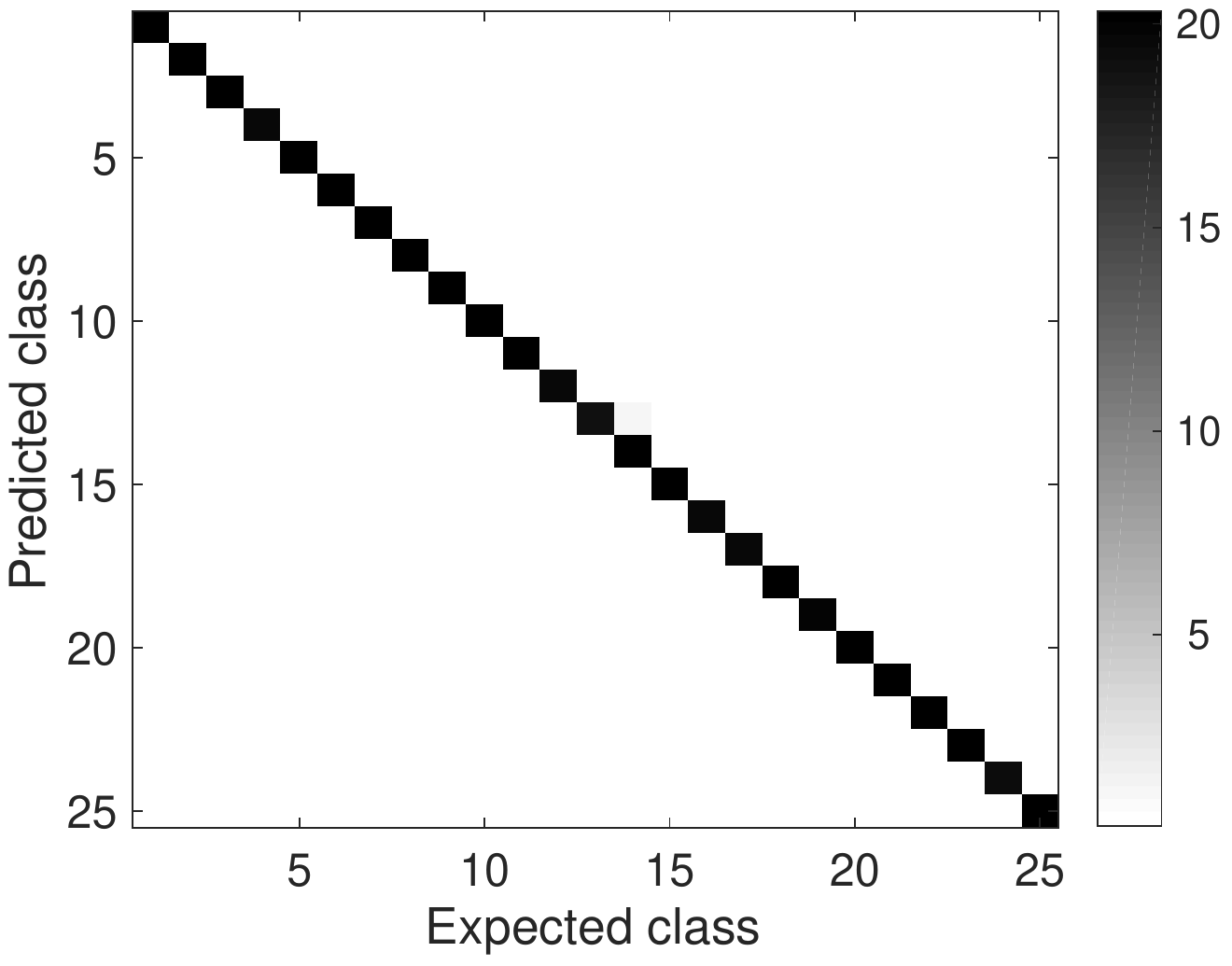}\\
			(c)\\
		\end{tabular}
	\end{tabular}
	\caption{Confusion matrices. (a) KTHTIPS-2b. (b) UIUC. (c) UMD.}
	\label{fig:CM1}
\end{figure}

Figure \ref{fig:CM1} shows the confusion matrices for the benchmark databases. Box counting descriptors were used to generate these pictures. Such representations essentially confirm the accuracies in Table \ref{tab:indiv}, but they also provide information regarding the accuracy in each class, what opens possibility for a more elaborate analysis on the classification outcomes. In Figure \ref{fig:CM1} (a) we see KTHTIPS-2b with lower accuracy in classes 3 (``corduroy'') and 5 (``cotton''). These are actually materials frequently confused as they are types of clothing fabrics and possess similar texture patterns. UIUC, on the other hand, yields a nearly perfect classification result, with some significant misclassification only in class 8 - granite -  (confused with 18 - carpet). Despite being different materials, they are both characterized by a granular appearance, which poses some difficulties for the automatic discrimination.

\subsection{Identification of Plant Species}

Table \ref{tab:SRdatabase_plant} lists the accuracy of the fractal-LBP descriptors in 1200Tex database \cite{CMB09}, compared with a few state-of-the-art results published in the literature. This is a set of images of plant leaves for 20 Brazilian species collected \textit{in vivo}. For each species 20 samples were collected, washed, aligned with the vertical axis and photographed by a scanner. The image of each sample was split into 3 non-overlapping windows with size $128\times 128$. Those windows were extracted from regions of the leaf presenting less texture variance and were converted into gray scale images, resulting in a database with 1200 images. From each species 30 images were randomly selected for training and the remaining images for testing. This procedure was repeated 10 times, which allowed the computation of the average accuracy and respective deviation (in Table \ref{tab:indiv} and Table \ref{tab:comb}). 
\begin{table}
	\centering
	\caption{Accuracy of the fractal-LBP descriptors compared with other results in the identification of plant species.}
	\label{tab:SRdatabase_plant}	
	\begin{tabular}{cc}
		\hline
		Method & Accuracy (\%)\\
		\hline
		LBPV \cite{GZZ10} & 70.8\\
		Network diffusion \cite{GSFB16} & 75.8\\
		FC-CNN VGGM \cite{CMKV16} & 78.0\\		
		Gabor \cite{CMB09} & 84.0\\
		FC-CNN VGGVD \cite{CMKV16} & 84.2\\
		Schroedinger \cite{FB17} & 85.3\\		
		SIFT + BoVW \cite{CMKMV14} & 86.0\\		
		\hline
		Proposed & 86.3\\
		\hline
	\end{tabular}
\end{table}

\begin{figure}
	\centering
	\includegraphics[width=.45\textwidth]{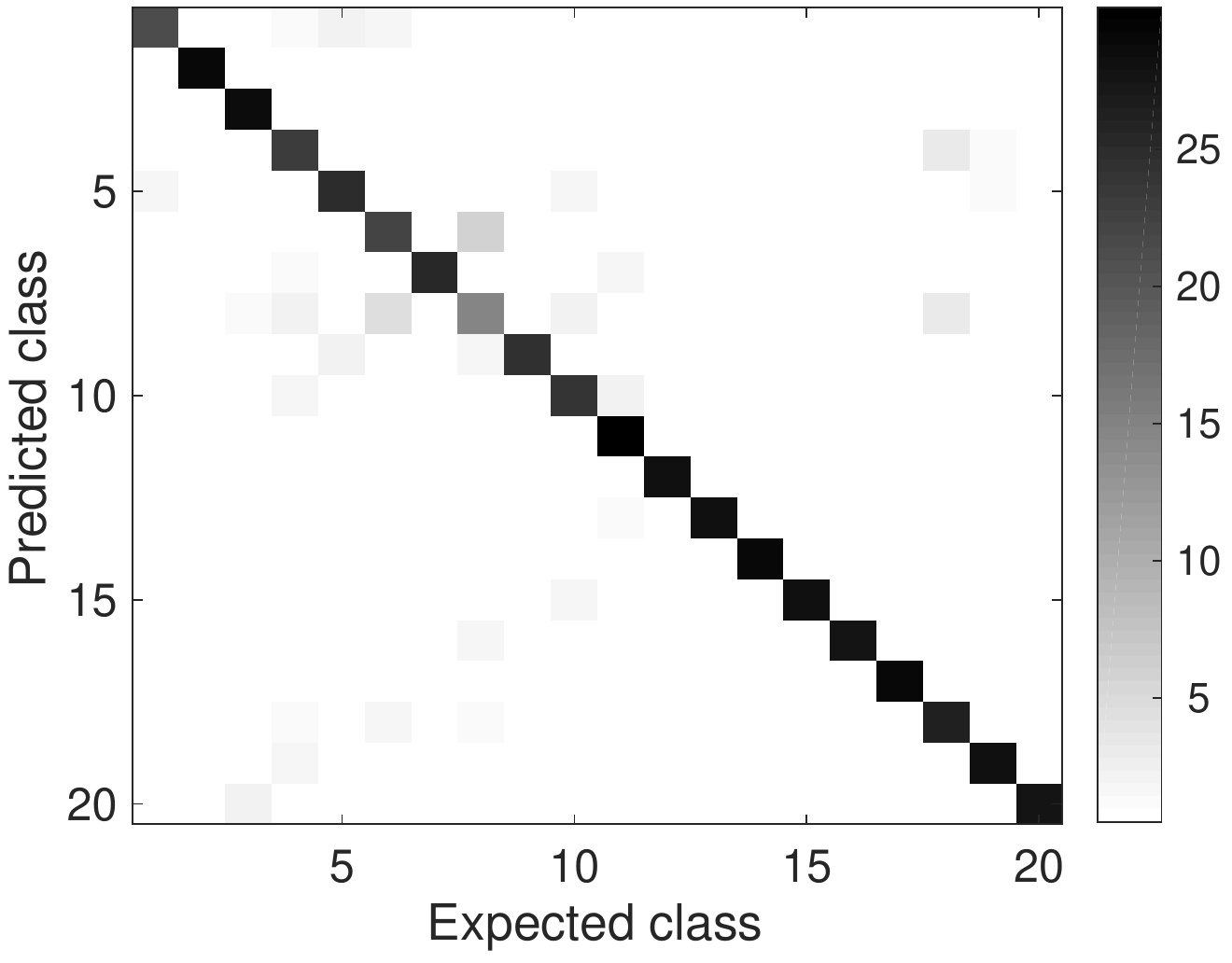}
	\caption{Confusion matrix for the plant database.}
	\label{fig:CM2}
\end{figure}

Figure \ref{fig:CM2} complements Table \ref{tab:SRdatabase_plant} by exhibiting the confusion matrix for the proposed method (box counting features). Generally speaking, the accuracies in all classes are high and the most critical situation occurs in class 8, which is confused for example with class 6. These correspond to samples with similar textures, especially with regards to the arrangement of nervures and leaf microtexture, which are prominent elements for the process of distinguishing among samples from different species.

Generally speaking, fractal descriptors still demonstrate competitiveness when compared with several state-of-the-art approaches for texture recognition. It is well known that characteristics intrinsic to the way that these materials are formed in nature contribute to this relation to a significant extent. This can be even more easily observed in data sets of ``pure'' textures, like UIUC and UMD, as well as in practical problems where such types of images naturally arise. This is the case in many biological applications and here is illustrated by the foliar surface images. The results encourage more research on this topic at the same time that it presents fractal descriptors as an alternative that should be verified in practical problems, as they can achieve competitive performance, for example, without requiring large amounts of training data and they also provide more natural interpretation for the obtained results as fractal sets have been classically associated with a mathematical model of nature structures.  

\section{Conclusions}

Here we developed a combination of local binary patterns with numerical estimates of fractal dimension. More specifically, we compute the dimension of the LBP codes thresholded at different levels to compose the image feature vector.

The performance of our proposal was assessed in the classification of benchmark databases typically used in the literature. We also employed such descriptors in a practical problem with significant importance in botany and related areas, namely, the identification of species of Brazilian plants. In both cases our method obtained promising results, comparable to state-of-the-art results recently published in the literature.

The results presented here suggest that the combination of fractal geometry (and potentially other fractal measures) with a local encoding like LBP can be rather useful to represent all the rich information conveyed by a texture image.

\section*{Conflict of interest}

The authors declare that they have no conflict of interest.


\end{document}